\documentclass{vgtc}                          

\ifpdf
  \pdfoutput=1\relax                   
  \usepackage{graphicx}                
  \DeclareGraphicsExtensions{.pdf,.png,.jpg,.jpeg} 
\else
  \usepackage{graphicx}                
  \DeclareGraphicsExtensions{.eps}     
\fi%

\graphicspath{{figures/}{pictures/}{images/}{./}} 

\usepackage{amsmath}                   
\usepackage{microtype}                 
\PassOptionsToPackage{warn}{textcomp}  
\usepackage{textcomp}                  
\usepackage{mathptmx}                  
\usepackage{times}                     
\usepackage{cite}                      
\usepackage{tabu}                      
\usepackage{booktabs}                  

 \usepackage{url}
\usepackage{hyperref}
\usepackage{amssymb}
\usepackage{verbatim}

\usepackage{enumitem}
\usepackage{lipsum}
 
\onlineid{0}

\vgtccategory{Research}

\title{An Explainable AI Approach to Large Language Model Assisted \\ Causal Model Auditing and Development}

\author{Yanming Zhang\thanks{e-mail: yanming.zhang@stonybrook.edu}\\ %
        \scriptsize Stony Brook University %
        \and Brette Fitzgibbon\thanks{e-mail: bconnolly6720@mycom.marin.edu}\\ %
     \scriptsize College of Marin  %
     \and Dino Garofolo\thanks{e-mail: dino.garofolo@stonybrook.edu}\\ %
     \scriptsize Stony Brook University %
     \and Akshith Kota\thanks{e-mail: akkota@cs.stonybrook.edu}\\ %
     \scriptsize Stony Brook University %
     \and Eric Papenhausen\thanks{e-mail: epapenha@akaikaeru.com}\\ %
     \scriptsize Akai Kaeru LLC %
     \vspace{10pt}
\and Klaus Mueller\thanks{e-mail: mueller@cs.stonybrook.edu}\\ %
     \parbox{1.4in}{\scriptsize \centering Stony Brook University \\ Akai Kaeru LLC}}

\teaser{
  \centering
\vspace{-5pt}
\includegraphics[width=1\linewidth]{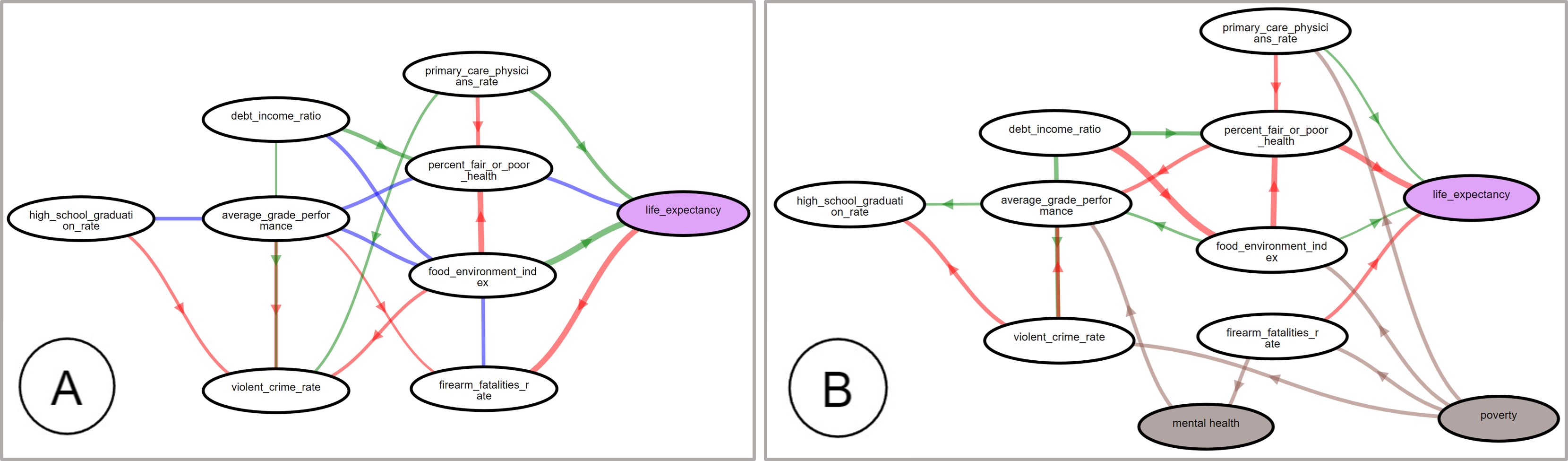}
\vspace{-15pt}
  \caption{Our ChatGPT-powered Causal Auditor. (A) \textbf{Initial Causal Model} generated by the PC algorithm, some edges are undirected, others might be incorrect (B) \textbf{ChatGPT Refined Model} with all audited edges and new latent variables. 
 }
  \label{fig:teaser}
}

\abstract{Causal networks are widely used in many fields, including epidemiology, social science, medicine, and engineering, to model the complex relationships between variables. While it can be convenient to algorithmically infer these models directly from observational data, the resulting networks are often plagued with erroneous edges. Auditing and correcting these networks may require domain expertise frequently unavailable to the analyst. We propose the use of large language models such as ChatGPT as an auditor for causal networks. Our method presents ChatGPT with a causal network, one edge at a time, to produce insights about edge directionality, possible confounders, and mediating variables. We ask ChatGPT to reflect on various aspects of each causal link and we then produce visualizations that summarize these viewpoints for the human analyst to direct the edge, gather more data, or test further hypotheses. We envision a system where large language models, automated causal inference, and the human analyst and domain expert work hand in hand as a team to derive holistic and comprehensive causal models for any given case scenario. This paper presents first results obtained with an emerging prototype.
} 

\begin{document}


\firstsection{Introduction}
\maketitle

Causal networks are widely used in various fields, such as epidemiology \cite{vandenbroucke2016causality}, healthcare \cite{glass2013causal}, biology \cite{dang2015reactionflow}, social sciences \cite{gerring2005causation}, etc. to model and understand complex systems. These networks typically consist of a set of nodes and edges, where the nodes represent the variables and the edges represent the causal relationships between them. As the accuracy and reliability of causal networks heavily depend on the assumptions and data used in the modeling process, it is essential to verify that the causal relationships represented in the network are consistent with the actual real-word relationships. 

One approach to assure the accuracy and reliability of causal networks is through auditing. Auditing involves evaluating the causal model to identify potential errors that may have been introduced during its development. Typically these activities require a qualified human domain expert who, however, might not be readily available to a model developer. As such, model auditing represents a significant bottleneck in causal model development. 

We propose a potential solution to mediate the scarcity of expert knowledge in model auditing activities, namely, automated chatbots, such as ChatGPT. ChatGPT is a large language model (LLM) based on transformer neural networks trained on a massive corpus of text data, including web pages, books, and other sources. It has already demonstrated impressive capabilities in generating natural language responses to various prompts, such as completing sentences, answering questions, and even generating coherent paragraphs. We propose that these capabilities will also make it an ideal candidate to help with auditing and even developing causal networks.

But as has been widely reported ChatGPT can fall victim to hallucinations, and asking it a simple question like “Does A cause B”, even when posed within a sophisticated prompt \cite{kiciman2023causal}, can result in erroneous statements. To lower the probability of false conclusions we have built a prompt template that asks the causal question in all possible polarities. We hypothesize that asking the same fundamental question from multiple angles can elevate confidence in the chosen relation and also expose conflicts associated with it.

Asking causal questions in multiple ways can also illuminate the broader context associated with this relation. In its justifications ChatGPT typically mentions confounder and mediator variables that surround the queried relation, but we can also specifically ask for it. 
Any new domain knowledge gained from this information can then prompt targeted data collection and causal model refinement, or even splitting the model into several contextual models. 

A downside of this general approach is that it generates a substantial amount of data, and to serve the mission of explainable AI the data are meant for human consumption to foster insight, confidence, and trust into the causal model. To aid this mission, we have devised a set of dedicated visualizations to make it easier for a human analyst to grasp the essence of this LLM-generated data. This paper reports on first studies we have conducted with this approach. 




In the following, Sections 2 and 3 discuss background and related work, Sections 4 and 5 describe our methodology and present further examples and results. Finally, Section 6 offers a conclusion.

\section{Background}

There are several strategies by which a causal directed acyclic graph (DAG) can be developed. The most rigorous one is to construct it from first principles via randomized controlled trials. This, however, is often not feasible due to cost, ethical, or practical constraints. It is also difficult to scale and restrictive on the number of researchers who can conduct such studies. A more scalable way is to \textit{discover} a causal DAG from observational data. There are essentially two popular strategies. One is to enforce the constraint that two statistically independent variables are not causally linked and then run a series of \textit{conditional independence tests} to construct a compliant DAG. Popular algorithms here are the PC \cite{spirtes2000causation} algorithm and the Fast Causal Inference (FCI) algorithm \cite{spirtes2001anytime}. Another way is to greedily explore the space of possible DAGs \cite{chickering2002optimal}; one may begin with an empty graph and iteratively add directed edges to maximize a model fitness measure such as the Bayesian Information Criterion (BIC) \cite{burnham2004multimodel, schwarz1978estimating}. 

\subsection{Shortfalls of Automated Causal Discovery}
Causal discovery entails four common assumptions: (1) the causal structure can be represented by an acyclic DAG, (2) all nodes are Markov-independent from non-descendants when conditioned on their parents, (3) the DAG is faithful to the underlying conditional independences, and (4) the DAG is sufficient in that there is no pair of nodes with a common external cause. Unfortunately these conditions are rarely entirely fulfilled, mostly due to selection/sampling bias in the observational data. While the probability of getting a (partially) incorrect DAG can be made arbitrarily small by using
enough data, it remains undefined how much data is actually needed \cite{Shalizi-chapter25, robins2003uniform}.   

The manifestation of the problem is the discovery algorithm's sensitivity to errors -- specifically, errors in the orientation of edges. If one edge is oriented incorrectly (e.g. the DAG states that \textit{A causes B}, when in reality \textit{B causes A}) then it may lead to a cycle. Since causal DAGs must be acyclic, the orientation of subsequent, correctly oriented edges, may be reversed. In this way, errors in the orientation of edges tend to propagate throughout the DAG. While there has been recent work \cite{rohekar2021iterative} that seeks to deal with this problem, progress so far has been limited. We therefore believe that the time is right to look for other solutions -- the vast causal knowledge stored in text documents. While extracting causal relations from text documents is not new \cite{yang2022survey}, thanks to ChatGPT they can now be conveniently unearthed with a simple prompt and for many application domains.

\section{Related Work}
Early work on the visualization of causal relations used animated views \cite{ elmqvist2003causality, kadaba2007visualizing}, while more recent research by Wang and Mueller \cite{wang2015visual} focused on visualizing causal relationships in form of Pearl’s DAG representation, with added support for analysts to interactively audit the DAG and gauge model quality with the BIC score. Their method used the PC algorithm to create the initial DAG. Conversely, Xie et al. \cite{xie2020visual} used a greedy score-based method, F-GES \cite{ ramsey2017million} to create the DAG and added mechanisms to enable overdraw-free graph layouts. The explanatory nature of causal DAGs readily affords What-if analyses. Here, Xie et al. took a cohort-based approach, while Hoque and Mueller \cite{hoque2021outcome} allowed individuals to gain insight on how the various factors causally affected the outcome of an automated decision system and at what sensitivity. Guo et al. \cite{guo2023causalvis} used the Rubin model to allow analysts to assess the potential effects of treatments on population subgroups, while Wang and Mueller \cite{wang2017visual} created DAGs for the various subgroups to explain their outcomes. Along similar lines, Yan et al. \cite{yan2020silva} and Ghai and Mueller \cite{ghai2022d} used DAGs to expose fairness issues in data used for model training data whereby the latter system also allowed users to mitigate these issues. Other researchers focused more on temporal aspects of causality \cite{wang2022domino, jin2020visual} also in text narratives \cite{choudhry2020once}, while yet others  \cite{kale2021causal, yen2019exploratory, xiong2019illusion} studied the perception of causality from a visual analyst’s point of view. Tookits for causal inference have also become available \cite{guo2021vaine, shimoni2019evaluation}. 

The literature on LLM-assisted causal network learning is rapidly emerging. The earliest documented attempt of using LLMs (here GPT-3) for causal analytics was published by Long et al.  \cite{long2023langcaus}. However, this was a rather preliminary study which focused on the structure of the prompt to reveal insight on the absence or presence of a directed edge. More recently, Kıcıman et al. \cite{kıcıman2023causal} studied the subject far more extensively, They developed a set of elaborate prompts for GTP-3.5 and GPT-4 that would force yes/no answers to standard causal queries. They showed excellent success rates for benchmark datasets where the causal truth was known but they did not attempt to use the GPT-output to acquire additional causal and contextual knowledge. They, and other papers that followed since then (e.g. \cite{nam2023show,long2023causal,gao2023chatgpt}), also did not attempt to visualize the acquired information in the realm of explainable and trustworthy AI. 








\begin{figure}[b!]
\centering
\includegraphics[width=\linewidth]
{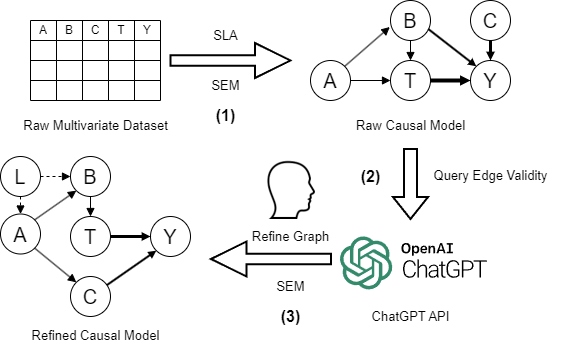}
\vspace{-15pt}
\caption{Workflow of our ChatGPT-powered Causal Auditor. (1) Algorithmic discovery of the initial (raw) causal model. (2) Query-driven ChatGPT-based edge commentary. (3) Analyst-initiated model refinement informed by the outcomes of steps 1 and 2.}
\label{fig:workflow}
\end{figure}

\section{Methodology}
Fig. \ref{fig:workflow} illustrates the workflow of our method. Step 1 computes an initial causal model from the user-provided data. This model is displayed as a DAG in a suitable visual causal analytics interface, such as \cite{wang2017visual, xie2020visual}. This model can then be refined based on the outcome of a ChatGPT prompt with regards to a certain relation indicated by the user. Apart from listing the outcomes of the prompts in form of text, we also provide visualizations that seek to summarize the gist of these responses for the user. 


We investigate causality at two levels. The first level seeks to identify whether a direct causal relation actually exists, while the second level dives deeper and checks for mediators and confounders. We could have used the slightly weaker term 'covariate' over 'confounder' but we found using 'confounder' produced richer results.

\subsection{Prompting for Direct Causal Relationships}

The prompts we use are according to the following template:

\vspace{5pt}

\noindent \textbf{Prompt:} On a scale from 1 to 4, where 4 represents strong or most likely, rate the cause-and-effect relationship: higher/lower {A/B} causes higher/lower {B/A}.

\vspace{5pt}

\noindent This generates 10 distinct prompts, 5 each for A and B taking opposite roles, and within each of these two sets there are 4 combinations of A and B being (higher, lower) plus one relation that just asks this for a general case. For example, we might have: 

\vspace{5pt}

\noindent \textbf{Prompt:} On a scale from 1 to 4, where 4 represents strong or most likely, rate the cause-and-effect relationship: \textit{higher percent fair or poor health} causes \textit{higher life expectancy}.

\vspace{5pt}

\noindent The below is an example for the general prompt:

\vspace{5pt}

\noindent\textbf{Prompt:}  On a scale from 1 to 4, where 4 represents strong or most likely, rate the cause-and-effect relationship: changing \textit{percent fair or poor health } causes a change in \textit{life expectancy}.

\vspace{5pt}

\noindent This prompt resulted in the following response using GPT-4 (some part are redacted for brevity):   

\vspace{5pt}

\noindent \textbf{Response}: .... I would rate the cause-and-effect relationship between the percentage of the population in fair or poor health and changes in life expectancy as a 4. This rating is based on the understanding that life expectancy is heavily influenced by the overall health status of a population. If a large percentage of the population is in fair or poor health, it's likely to negatively affect the average life expectancy. Poor health status could be linked to a range of factors including chronic diseases, infectious diseases, poor mental health, and lifestyle-related health conditions. All these factors can shorten individual lifespans, leading to a lower average life expectancy for the population as a whole. However, it's important to note that this is a general trend and doesn't necessarily hold in every specific case. Other factors can also impact life expectancy, such as healthcare access and quality, socioeconomic status, environmental conditions, and more...

\subsection{Prompting for Confounders and Mediators}

The prompt template we developed here also distinguishes among the 4 combinations that explore the effects of higher and lower levels plus one relation that just asks this for the general case. In the following we explain this template using the variables \textit{food environment index (FEI)} and \textit{violent crime rate} as an example.

\vspace{5pt}

\noindent \textbf{Prompt:} On a scale from 1 to 4, 4 represents strong or most likely, rate the cause-and-effect relationship 'For a county, lower \textit{food environment index} causes higher \textit{violent crime rate}'. Make a concise list of mediators in that relation and assign strengths to them (weak, medium, strong). Also make a concise list of confounders in that relation and assign strengths to them (weak, medium, strong).

\vspace{5pt}

\noindent We found that by specifying 'county' we obtained a much more concrete list that made better references to the data at hand. 

\vspace{5pt}

\noindent This prompt resulted in the following response using GPT-4. After noting that the causal relation can be rated a 2, indicating a "somewhat plausible link" it wrote (some part are redacted for brevity):

\vspace{5pt}

\noindent \textbf{Response}: GPT-4 suggested the following three mediators:
\vspace{-5pt}
\begin{itemize} [leftmargin=*]
    
\item \textbf{Poverty (Strong)}: Poorer food environments (PFE) often correlate with higher poverty levels, which in turn can lead to increased crime due to socioeconomic frustrations, lack of opportunities, or the need to commit crime to survive. 
\vspace{-5pt}
\item \textbf{Educational Attainment (Medium)}: PFE may align with areas of lower educational attainment. Lower levels of education can contribute to higher crime rates due to fewer opportunities for gainful employment and lack of knowledge about alternatives.
\vspace{-7pt}
\item \textbf{Health Outcomes (Weak)}: PFE can lead to negative health outcomes, which could potentially increase crime rates indirectly. Poor health may lead to higher levels of stress in the community, which can exacerbate social issues that contribute to crime. 
\end{itemize}

\noindent GPT-4 also suggested the following three confounders. 
\begin{itemize} [leftmargin=*]
\vspace{-7pt}
\item \textbf{Socioeconomic Status (Strong)}: Lower socioeconomic status often coincides with both PFE and higher crime rates. This may make it seem like PFE cause higher crime rates when the real culprit is socioeconomic status. 
\vspace{-7pt}
\item \textbf{Urban vs Rural Setting (Medium)}: Urban areas tend to have both lower FEI and higher crime rates due to higher population density, greater economic disparities, and other factors. This can confound the relation between FEI and crime rates. 
\vspace{-7pt}
\item \textbf{Public Policy (Weak)}: Public policies can influence both the food environment (through regulations, subsidies, etc.) and crime rates (through law enforcement funding, criminal justice policies, etc.), acting as a confounding factor.
\end{itemize}

\noindent Last, GPT-4 reminded us that "correlation does not imply causation, and while these relationships can exist, they do not guarantee that a lower FEI will always lead to a higher crime rate. Other factors, such as strong community leadership, effective policing, and active non-profit organizations, can also have significant impacts on crime rates". This message hints on an important truth which applies to almost every causal relationship, even those with higher ratings. There are almost always mediators and covariates that play a role. They can contextualize the causal relation to certain sub-populations and they can also offer effective means for intervention. This has been a strong motivator for the visual interface we have designed. 

\subsection{Visualizing the GPT-4 Generated Text Responses }

Once that one acknowledges that a binary score alone is not all that counts to judge a causal relation, the problem now becomes that there is large amount of textual information that GPT-4 generates. Even when instructed to just summarize the findings, this wealth of information can be overwhelming to a general user. In the following we describe a few dedicated visualizations we have designed to make browsing this information easier. 

The first step in deriving a visualization is to identify and retrieve the key information from the text responses. For this we have made use of standard NLP techniques available in libraries like Spacy\footnote {https://spacy.io} and Allen NLP\footnote{https://allenai.org/allennlp}. Specifically. we have used these libraries to identify the various variables, mediators, and covariates, their importance ratings and their interrelations in the GPT-4 generated text responses. 

In the following we describe the various charts we have created. 


\vspace{5pt}

\subsection{The Causal Debate Chart}

\begin{figure}[t!]
\centering
\includegraphics[width=0.9\linewidth]{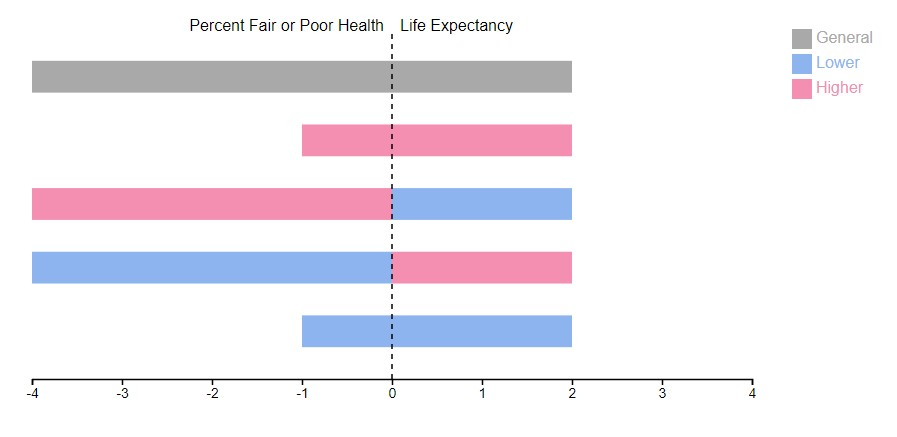}
\vspace{-5pt}
\caption{Causal Debate Chart for the relation \textit{Percent Fair or Poor Health - Life Expectancy}, presenting an overwhelming belief that the former is the cause of the latter.}
\label{fig:FH-LE}
\end{figure}

Fig. \ref{fig:FH-LE} shows our summary visualization that contrasts the numerical outcomes of these 10 prompts. We call it \textit{Causal Debate Chart} since it visually argues the strength of one variable being the cause of the other. The chart is a bidirectional bar chart where each side is headed by one of the two relation variables. In  this case the left side is \textit{Percent Fair or Poor Health} and the right side is \textit{Life Expectancy}. The x-axis is the score assigned by ChatGPT and the length of each bar is mapped to that score. The grey bars are for the general prompt while the other bars are colored in red if the cause was a higher or increasing level of the variable or in blue if the cause was a lower or decreasing level (see color legend on the top right). 

Let's evaluate the chart. We observe that for the first (grey) set of bars \textit{Percent Fair or Poor Health} has a substantially longer bar (level 4) than \textit{Life Expectancy} which has level 2 (level 2 is a rather weak cause in GPT-4 semantics). It means that the former wins the causal debate -- it has causal dominance. \textit{Percent Fair or Poor Health} seems to be a general cause of \textit{Life Expectancy}

Now let us check out the other bars which play out specific level studies. There we see if ChatGPT is consistent in its logic (as opposed to hallucinating). We confirm that also here we see that high or increasing \textit{Percent Fair or Poor Health} leads to low or decreasing \textit{Life Expectancy}. The same is true for the opposite. Finally, the other two sets have low bars on both sides, which is to be expected if the relation as indicated by the other bars is taken as true. In some sense the Causal Debate Chart of Fig. \ref{fig:FH-LE} is a prime example for what we would expect from a steadfast causal relation.

\begin{figure}[t!]
\centering
\includegraphics[width=0.9\linewidth]{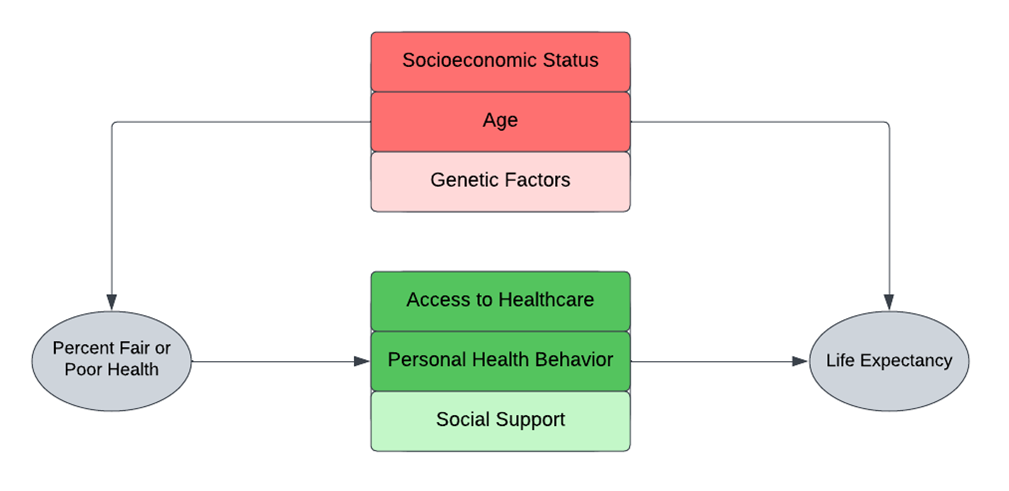}
\caption{Causal Relation Environment Chart for the relation \textit{Percent Fair or Poor Health - Life Expectancy}. The intensity of red and green encodes the strength of the mediators and covariates (weak, medium, strong), and the color of the cause and effect variables are alike those in Fig. \ref{fig:FH-LE}; in this specific case they are grey.}
\label{fig:PFGH-CR}
\end{figure}

\subsection{The Causal Relation Environment Chart}
The Causal Relation Environment Chart is a diagram that shows the complete causal relation, consisting of the two main relational variables along with a list of mediators and confounders/covariates. An example for the general \textit{Percent Fair or Poor Health (PFPH) - Life Expectancy (LE)} relation is shown in Fig. \ref{fig:PFGH-CR}. 

\begin{figure}[t!]
\centering
\includegraphics[width=0.9\linewidth]{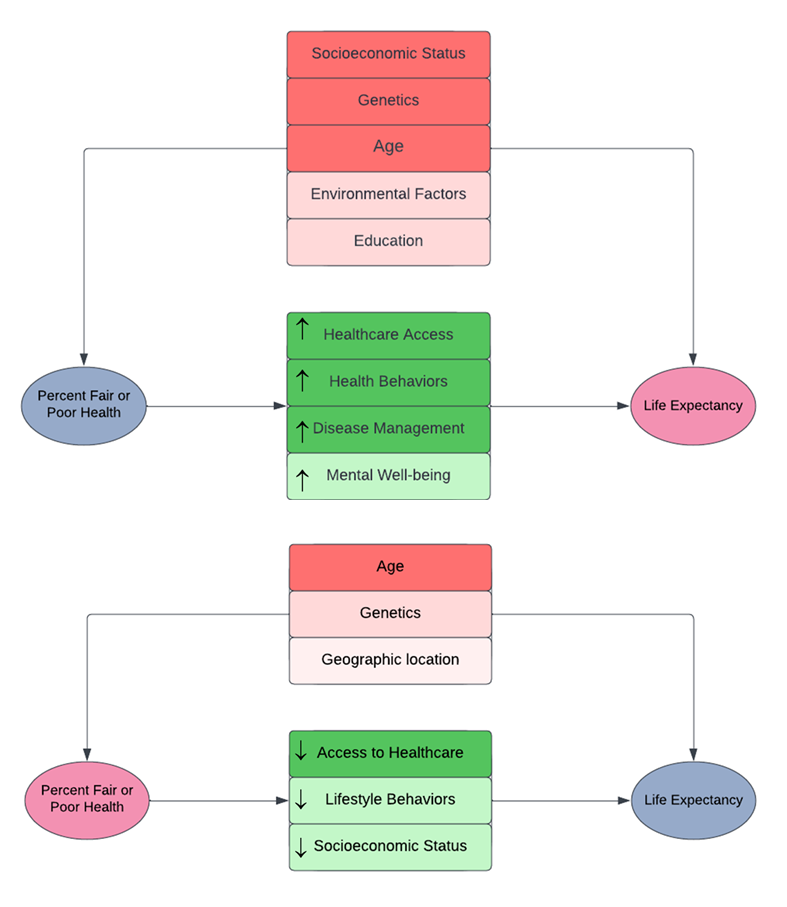}
\caption{Causal Relation Environment Chart for two level combinations of the relation \textit{Percent Fair or Poor Health - Life Expectancy}. The up and down arrows show the appropriate signs of the mediators.}
\label{fig:PH-LE 1}
\end{figure}

It turns out that the more focused (low, high) relations, even those with low causal score, have quite similar mediators and covariates. But they vary in the sign. For instance to go from \textit{low PFPH} to \textit{high LE} positive levels of the mediators are cited, such as good access to healthcare and good health habits, while to go from \textit{high PF-PH} to \textit{low LE} the cited mediators are usually the opposite, like limited access to health care and poor health habits. These two cases are shown in Fig. \ref{fig:PH-LE 1} where the up and down arrows indicate the positive and negative levels, respectively. 

\begin{figure}[t!]
\centering
\includegraphics[width=0.9\linewidth]{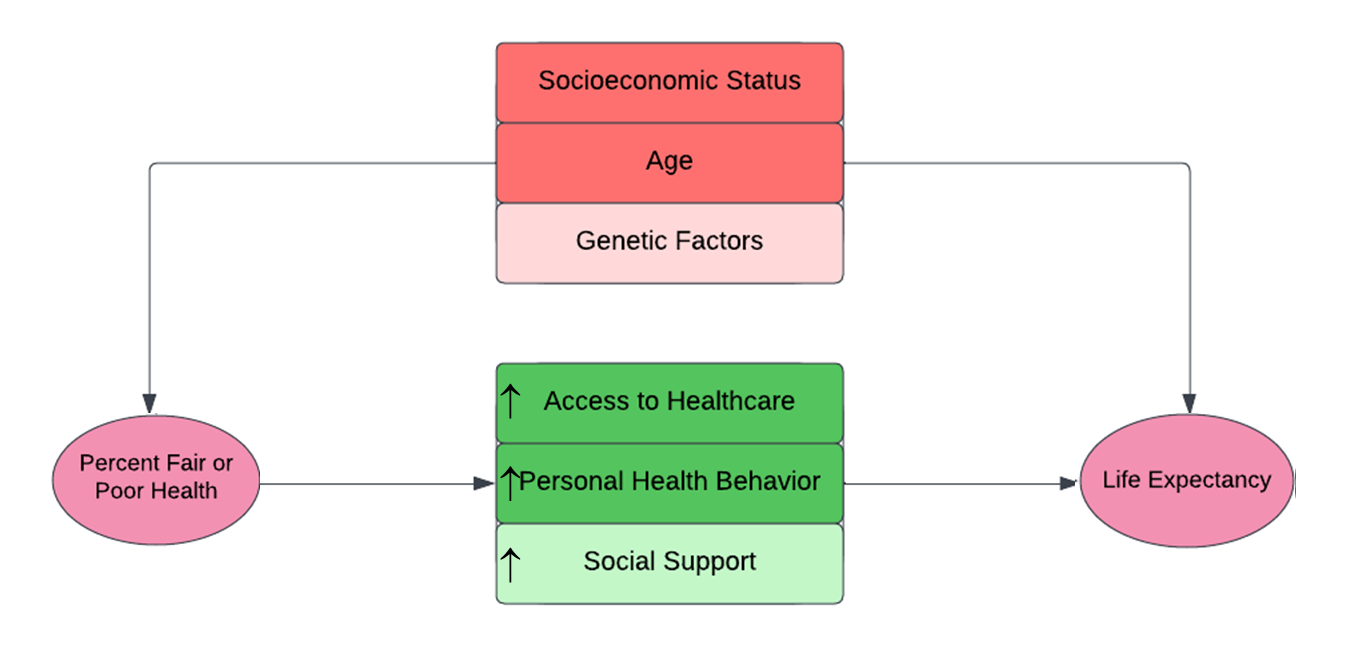}
\caption{Causal Relation Environment Chart for an improbable level combination of the relation \textit{Percent Fair or Poor Health - Life Expectancy}, namely one where both variables have positive level. The up arrows in the mediators show how this improbable combination might be achieved, in form of interventions on the mediators in the direction of the arrows. }
\label{fig:HPH-HLE}
\end{figure}

In this context, it is interesting to examine the relation \textit{high PFPH} to \textit{high LE} which is deemed improbable (see Fig. \ref{fig:FH-LE}). For this case, GPT-4 correctly responds that this relation is "counterintuitive" but then goes along with it and "considers it as a hypothetical scenario". Following this prelude, it suggests that the mediators need to be "improved" or "strong" to achieve the desired \textit{high LE}. In essence, these are all potential interventions a policymaker could undertake to fix an undesirable \textit{high PFPH} to \textit{low LE} and make the improbable \textit{high PFPH} to \textit{high LE} a reality. Fig. \ref{fig:HPH-HLE} visualizes the Causal Relation Environment Chart for this case. However, given its semantics we might rename it to \textit{Causal Intervention Chart}. 

We have observed that for confounders the situation is not as clear-cut and we intend to study this further in future work. 

\subsection{The Confounder/Mediator Chart}

\begin{figure}[t!]
\centering
\includegraphics[width=0.9\linewidth]{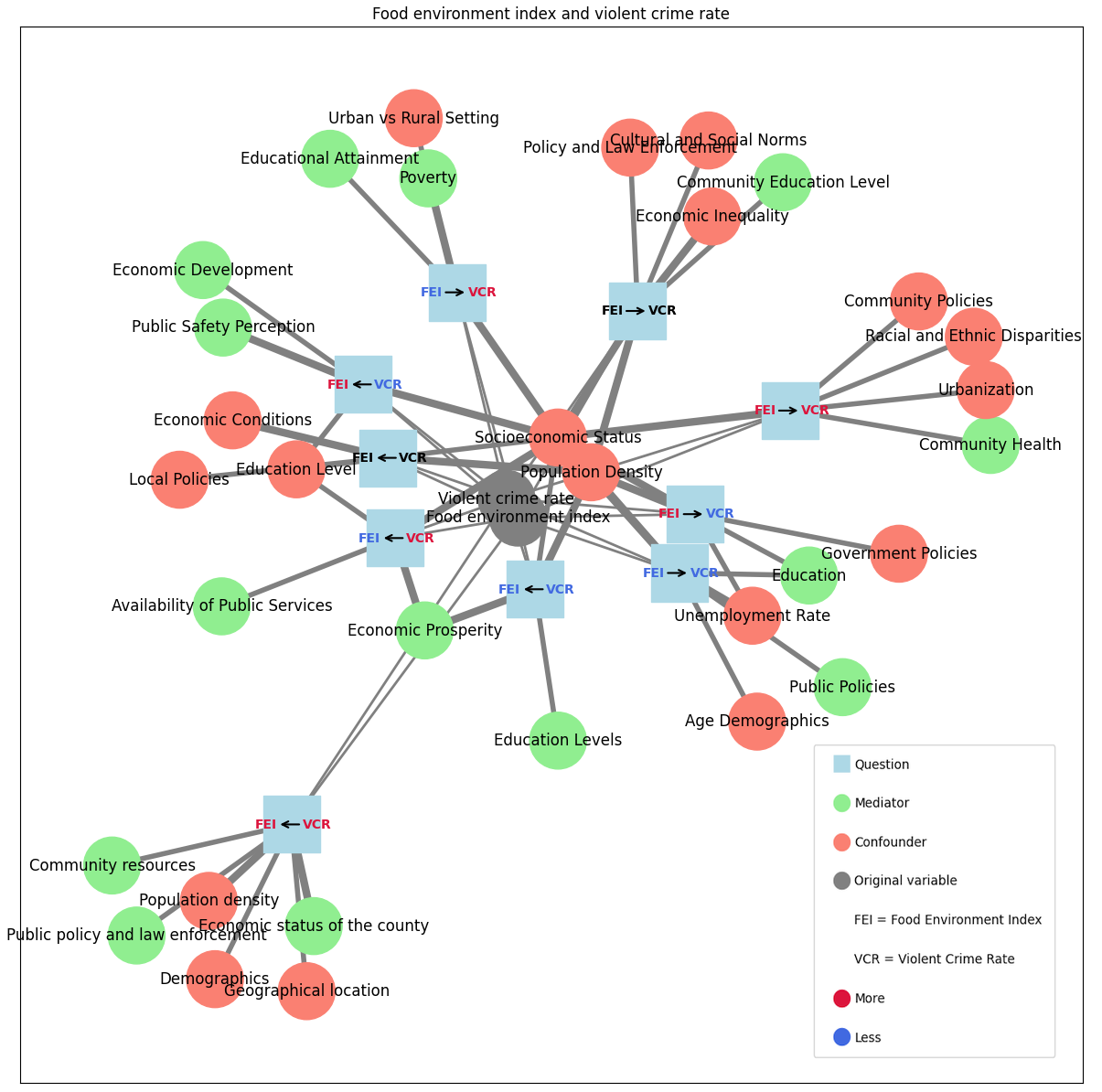}
\caption{Confounder/Mediator Chart for the relation \textit{Food Environment Index (FEI)} and \textit{Violent Crime Rate (VCR)}. The legend on the bottom right explains the various symbols and colors used.}
\label{fig:CMG}
\end{figure}

The Causal Debate Chart can quickly communicate the numerical outcomes of the 10 different causal level combinations and the Causal Relation Environment Chart can focus on the confounders and mediators of one such causal relation at a time. To give a full overview we have designed a summary chart we call the \textit{Confounder/Mediator Chart}. Fig. \ref{fig:CMG} shows an example for the causal relation \textit{Food Environment Index (FEI)} and \textit{Violent Crime Rate (VCR)},

This chart represents each causal relation and level combination as a light-blue box (called a question) where the color of the inscribed text denotes the level of each of the two causal variables. The identified mediators (green disks) and confounders (red disks) emerge from their respective questions where the width of the connecting edge denotes the influential strength as identified by GPT-4. 

A first observations we make in this chart is that the most common confounders are Socioeconomic Status and Population Density which are plotted at the center of the graph. These are the common themes under which this causal relation "lives". 

By looking at and around the individual relations one can make out more specific scenarios only valid for one or a subset of relations. One of these is the relation on the right, high FEI and high VCR. Why could there be quality food but still high crime rate? The chart shows that racial and ethnic disparities (which usually occur in urban environments) and community policies can set the contextual stage for this (as confounders). 

Interesting is also to investigate how high VCR can lead to low and high FEI, respectively. For the former, we see a mediator \textit{Availabilty of Public Services}, or better, the lack thereof. Conversely, for the latter, we see mediators like \textit{Public Policy and Law Enforcement} that ensure that FEI remains high despite high VCR.

\section{Further Examples and Results\label{useage}}
We have prepared three further examples to demonstrate our method. For brevity we only show the respective Causal Debate Charts. 

\begin{figure}[t!]
\centering
\includegraphics[width=0.9\linewidth]{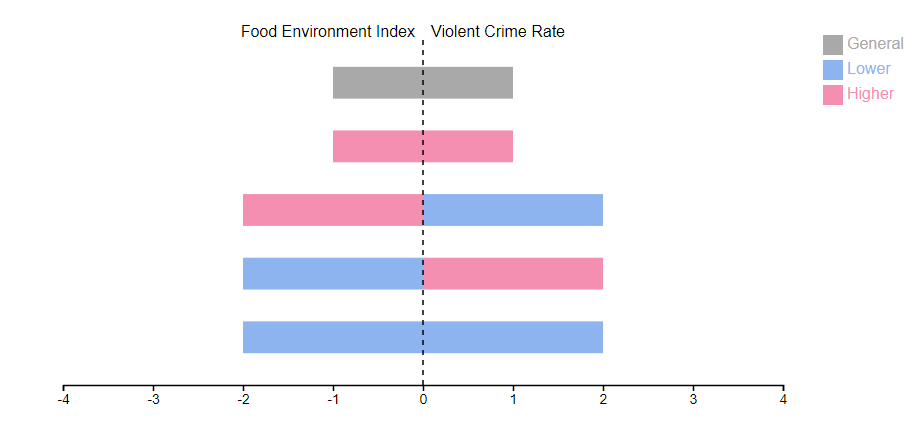}
\caption{Causal Debate Chart for the relation \textit{Food Environment Index - Violent Crime Rate}. There does not seem a direct causal relation between these two variables for any polarity.} 
\label{fig:FEI-VCR}
\end{figure}

\textbf{Food Environment Index and Violent Crime Rate.} Fig. \ref{fig:FEI-VCR} shows the Causal Debate Chart for the relation between \textit{Food Environment Index (FEI)} and \textit{Violent Crime Rate (VCR)}. We observe that there only weak (level 1-2) support for a direct causal relation. 
But even though the direct causal relation is not strong, the Causal Debate Chart does not indicate that it is an absolute zero. And indeed, the Confounder/Mediator Chart in Fig. \ref{fig:CMG} brings up some interesting mediators as well as local contextual confounders which apply only for a subset of causal conditions. We have discussed some of these in the previous section.

\begin{figure}[t!]
\centering
\includegraphics[width=1\linewidth]{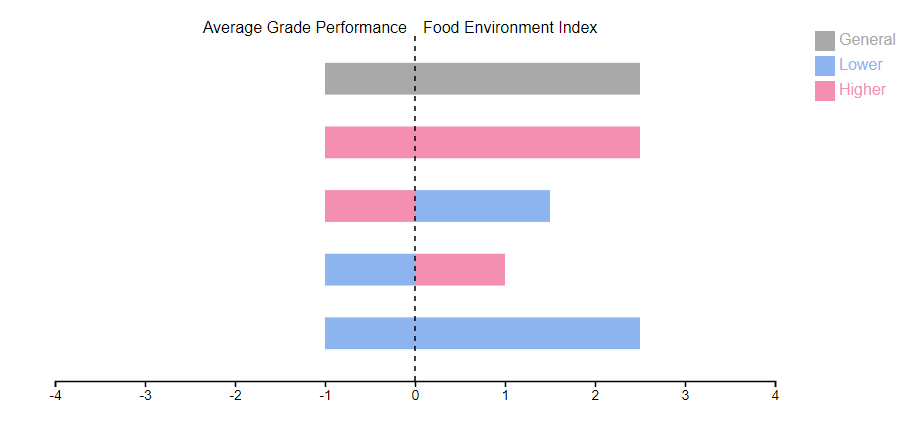}
\caption{Causal Debate Chart for the relation \textit{Average Grade Performance - Food Environment Index}. food environment index positively changes average grade performance, ChatGPT provides a medium score since a variety of factors makes the assessment complex; there is only a mild causation between these two attributes. } 
\label{fig:AGP-FEI}
\end{figure}

\textbf{Average Grade Performance and Food Environment Index} Fig. \ref{fig:AGP-FEI} shows the Causal Debate Chart for the relation between \textit{Average Grade Performance (AGP)} and \textit{Food Environment Index (FEI)}. There is a mild causal relation between the two, going from FEI to AGP. The chart provides coherent evidence for this. For FEI the grey bar is 2.5 times longer than the grey bar for AGP. And likewise, for both mixed (red-blue, high-low, opposite) relations the FEI bar is 2.5 or 1.5 times longer which is causally consistent, albeit not overly strong. Further analysis reveals that the relations is confounded by many other factors such as socioeconomic status, education level, cultural preferences, etc.

\begin{figure}[h!]
\centering
\includegraphics[width=1\linewidth]{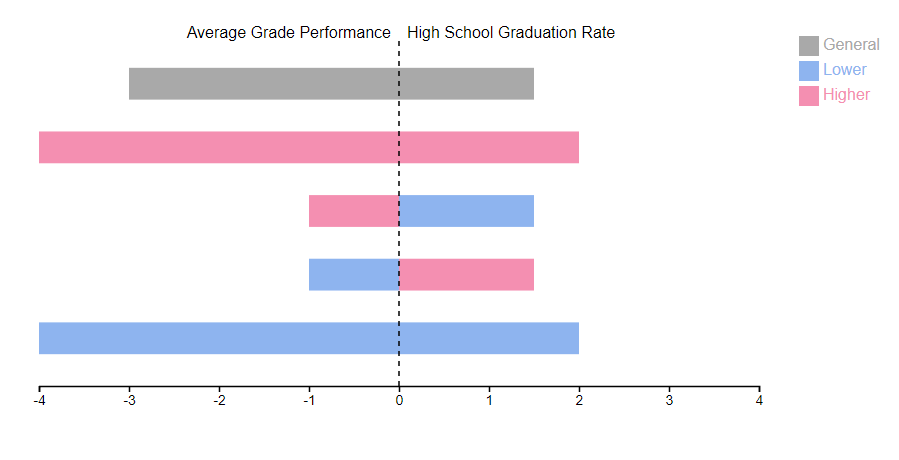}
\caption{Causal Debate Chart for the relation \textit{Average Grade Performance - High School Graduation Rate}. We observe that Average Grade Performance positively changes High School Graduation rate.} 
\label{fig:AGP-HGR}
\end{figure}

\textbf{Average Grade Performance and High School Graduation Rate} Fig. \ref{fig:AGP-HGR} shows the Causal Debate Chart for the relation between \textit{Average Grade Performance (AGP)} and \textit{High School Graduation Rate (HGR)}. There is a strong cause and effect relation among the two. This time there is positive causal relationship between the two variables, namely (high/low) AGP leading to (high/low) HGR.


\subsection{Evolved Causal Model}
We used the new insight gained from the GPT-4 based analysis to evolve the initial causal model we had computed via the PC-algorithm. The initial model is shown in Fig. ~\ref{fig:teaser}A, and the evolved model is shown in Fig.~\ref{fig:teaser}B. Since the model evolution is not the main focus of this paper -- the visualization of the GPT-4 text is -- we will not delve much into it and only present some results. 

\begin{figure}[h!]
\vspace{-5pt}
\centering
\includegraphics[width=\linewidth]{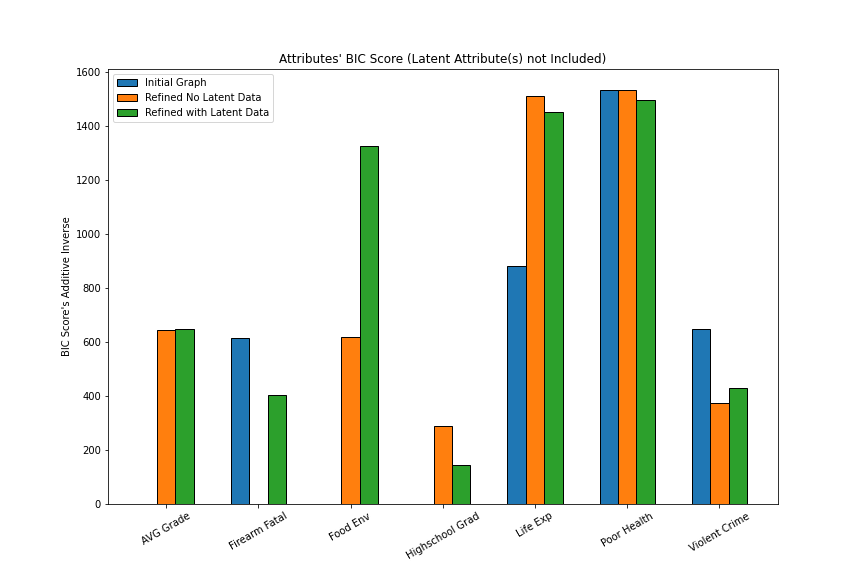}
\vspace{-25pt}
\caption{Attribute BIC scores before and after various stages of the audit, presenting the fit score for each attribute (higher is better). The blue bars are the scores for the initial model; the orange bars are the scores for the audited model but without adding the data for the newly identified mediators and confounders; the green bars are the bars after recomputing the model with the data of the newly introduced variables. For some variables this last update caused a small score reduction but the reduction was only minor.}

\vspace{-5pt}
\label{fig:attrbic}
\end{figure}
\begin{figure}[h!]
\vspace{-5pt}
\centering
\includegraphics[width=\linewidth]{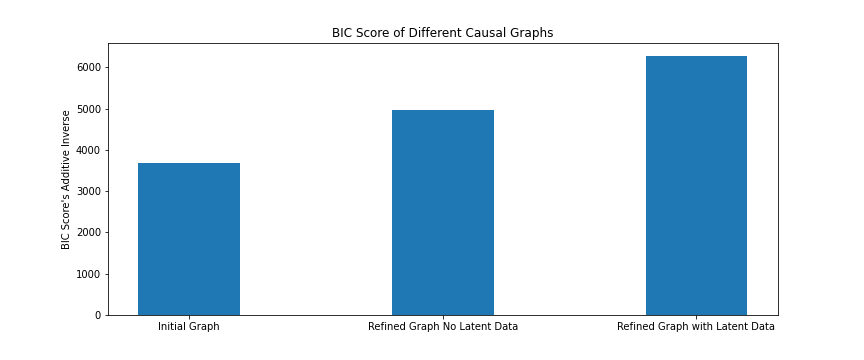}
\vspace{-25pt}
\caption{Overall graph BIC scores before and after various stages of the audit, presenting the fit score for each attribute (higher is better),}
\vspace{-5pt}
\label{fig:graphbic}
\end{figure}

In our initial experiment we used Poverty as a proxy for the various nuanced confounder variables we identified. 
As shown in Fig. \ref{fig:graphbic} and Fig. \ref{fig:attrbic}, the audition with GPT-4 increased the overall graph fit by 70.69\% (BIC score from -3673.4 to -6270.3), and for most attributes the BIC score improved.

\subsection{A Note on GPT-4 Accuracy}
We found that GPT-4 was significantly more accurate than GPT-3.5, which has been reported widely. Concretely, in our experiments GPT-4 produced the correct direction 103/110 times (94\%) and produced a number 109/110 times (99\%). For the "greater/lower A/B causes greater/lower B/A" queries, the 42 queries that proposed the inverse of the correct relationship (e.g., "greater average grade performance causes greater violent crime rate") had a range of 1 to 3 and a median of 1. The 68 queries that proposed the correct relationship, including "a change in A/B causes a change in B/A", had a range of 1 to 4 and a median of 2. Based on these observations we feel that GPT-4 is a valuable causal assistant but still requires human supervision.

\section{Conclusion}
We feel that the integration of ChatGPT as a Causal Auditor represents a significant stride forward in the causal inference field. By facilitating interactive model editing through easily accessible textual causal knowledge, this approach holds great promise in advancing our capacity to build better causal models and expand the frontiers of our understanding across various disciplines. 

To best utilize the vast knowledge GPT-4 can bring across we need techniques that allow humans to quickly get the gist of it and then delve in further. This is a typical "Overview first, zoom and filter, then details-on- demand" paradigm -- the visual information seeking mantra. We have presented a first step toward this premise. In the future we plan to add more information richness to our Causal Relation Environment Chart such that it can better capture the intricate information conveyed in the text. Also, the Confounder/Mediator Chart is still too cluttered and better layout algorithms are needed. 

It should be noted that the causal knowledge that can be extracted from the public facing version of ChatGPT is restricted by the resources that were used to train its large language model. So far we have not experienced any limitations but we have not yet tested an exhaustive variety of domains. If need arises an alternative might be to train one of the several open source large language models with a more specific corpus of text.   

We are currently developing an interactive interface by which users can click on a causal edge to verify its direction, contextualize it in terms of the population to which its relation applies, and refine it by the ChatGPT-discovered mediators, confounders, and colliders. This will then give way to an iterative process in which the causal model is continuously refined and where the causal edge weights are determined by additional data pulled from public or private repositories.  Once fully developed we will then study this interface with actual human users and observe how they use the system in their model auditing and development activities.

\section*{Acknowledgements}
This research was funded in part by a grant from the American Public Health Association (APHA) and a grant from the New York State Strategic Partnership for Industrial Resurgence (SPIR) program.



\bibliographystyle{abbrv-doi}

\bibliography{template}
\end{document}